\documentclass[twocolumn,10pt]{article}

\usepackage[square,numbers,sort&compress,comma]{natbib}
\usepackage{amsmath}
\usepackage{amssymb}
\usepackage{caption}
\usepackage{graphicx}
\usepackage{latexsym}
\usepackage{times}
\usepackage[pagewise]{lineno}
\usepackage{hyperref}
\usepackage{booktabs}
\usepackage{tabularx}
\usepackage{array}
\usepackage{color}

\renewenvironment{abstract}%
              {%
               \small%
               {\bfseries \abstractname}%
               \par%
               \vspace{10pt}%
              }%
              {}

\renewcommand\abstractname{Abstract}

\newcommand{\nomenclature}[1]{%
               \bgroup%
               \flushleft%
               \small\bf%
               #1%
               \par%
               \egroup%
              }

\renewcommand{\section}[1]{%
               \bgroup%
               \flushleft%
               \small\bf%
               \refstepcounter{section}%
               \arabic{section}. #1%
               \par%
               \egroup%
              }

\renewcommand{\subsection}[1]{%
               \bgroup%
               \flushleft%
               \small\em%
               \refstepcounter{subsection}%
               \arabic{section}.%
               \arabic{subsection}. #1%
               \par%
               \egroup%
              }

\renewcommand{\subsubsection}[1]{%
               \bgroup%
               \flushleft%
               \small\em%
               \refstepcounter{subsubsection}%
               \arabic{section}.%
               \arabic{subsection}.%
               \arabic{subsubsection}. #1%
               \par%
               \egroup%
              }

  \newcommand{\acknowledgement}[1]{%
               \bgroup%
               \flushleft%
               \small\bf%
               #1%
               \par%
               \egroup%
              }

  \newcommand{\sectionbib}[1]{%
               \bgroup%
               \flushleft%
               \small\bf%
               #1%
               \par%
               \egroup%
              }

\begin{document}
% -------------------------------------------------------------------- %

\small
\baselineskip 10pt

% -------------------------------------------------------------------- %
% --- Main Paper ---
% -------------------------------------------------------------------- %

\title{\LARGE \bf A unified foundational framework \\ for knowledge injection and evaluation of \\Large Language Models in Combustion Science}

\author{{\large Zonglin Yang$^{a}$, Runze Mao$^{a,*}$, Tianhao Wu$^{a}$, Han Li$^{a,b}$, QingGuo Zhou$^{b}$, and Zhi X. Chen$^{a,b,*}$}\\[10pt]
        {\footnotesize \em $^a$School of Mechanics and Engineering Science, Peking University, Beijing, 100871, PRChina}\\[-5pt]
        {\footnotesize \em $^b$AI for Science Institute, Beijing, 100080, China}}
\date{}

\twocolumn[\begin{@twocolumnfalse}
\maketitle
\rule{\textwidth}{0.5pt}
\vspace{-5pt}

\begin{abstract}
% General-purpose large language models (LLMs) excel at language-grounded reasoning and tool mediation, motivating their adaptation to scientific domains. 
To advance foundation Large Language Models (LLMs) for combustion science, this study presents the first end-to-end framework for developing domain-specialized models for the combustion community. The framework comprises an AI-ready multimodal knowledge base at the 3.5 billion-token scale, extracted from over 200{,}000 peer-reviewed articles, 8{,}000 theses and dissertations, and approximately 400{,}000 lines of combustion CFD code; a rigorous and largely automated evaluation benchmark (\textit{CombustionQA}, 436 questions across eight subfields); and a three-stage knowledge-injection pathway that progresses from lightweight retrieval-augmented generation (RAG) to knowledge-graph–enhanced retrieval and continued pretraining. We first quantitatively validate Stage 1 (naive RAG) and find a hard ceiling: standard RAG accuracy peaks at 60\%, far surpassing zero-shot performance ($\sim$ 23\%) yet well below the theoretical upper bound ($\sim$ 87\%). We further demonstrate that this stage's performance is severely constrained by context contamination. Consequently, building a domain foundation model requires structured knowledge graphs and continued pretraining (Stages 2 and 3).

\end{abstract}

\vspace{5pt}
\parbox{1.0\textwidth}{
}

\rule{\textwidth}{0.5pt}

\noindent *Corresponding author.
\vspace{5pt}
\end{@twocolumnfalse}]

% -------------------------------------------------------------------- %
% --- Section 1: Introduction ---
% -------------------------------------------------------------------- %

\section{Introduction\label{sec:introduction}}
\addvspace{10pt}

Since GPT-3 and its successors~\cite{Brown.2020}, Large Language Models (LLMs) have achieved striking progress in language-grounded reasoning and tool mediation~\cite{Luo.2025}. This achievement demonstrates a paradigm in which ingesting vast amounts of human domain data allows an AI system to internalize knowledge and acquire corresponding competencies~\cite{Song.2025}. Inspired by this, a natural idea has gained traction in scientific applications: with large-scale field corpora, a general-purpose LLM should be able to be adapted into a vertical scientist capable of domain reasoning and of orchestrating toolchains to carry out research autonomously~\cite{Chen.2024,Yue.2025,Dong.2025,Feng.2025}.  This hypothesis is highly compelling, yet the technical pathway for its realization remains open.

% are being applied to scientific research workflows \cite{Brown.2020, Luo.2025}, but in specialized domains like combustion science, they face challenges due to a lack of profound knowledge \cite{Sharma.2024, Dong.2025}. Therefore, a robust domain knowledge acquisition-injection framework is necessary \cite{Song.2025}.

Recently, researchers in the combustion community have begun exploring domain-specific LLMs, most commonly via retrieval-augmented generation (RAG)~\cite{Lewis.2020}, a lightweight strategy that injects curated external context into prompts without modifying model weights. For example, Sharma et al.~\cite{Sharma.2024} address oblique detonation by assembling a small knowledge base of about 100 papers and building a RAG pipeline for a general-purpose LLM such as ChatGPT; they report preliminary reliability on a limited test of three Q\&A pairs. Such explorations are valuable but remain at the proof-of-concept stage; their narrow scenario focus and limited domain coverage make hallucinations likely and hinder practical deployment. Therefore, we contend that what will ultimately serve the community is a domain foundation model that is comprehensively knowledgeable, continuously learnable, and dependable across applications, serving as a trustworthy authority in combustion science.

% Existing knowledge injection methods form a "spectrum" of resource requirements: at one end are data-driven methods like CPT and SFT \cite{Gururangan.2020, Ouyang.2022, Hu.2021}, which require massive resources; at the other end is Retrieval-Augmented Generation (RAG) \cite{Lewis.2020}, which, as a lightweight solution, has become the "preferred exploratory path" for domain applications.

% However, the true effectiveness of this "preferred path" remains largely unquantified. Despite pioneering explorations \cite{Sharma.2024}—for instance, Sharma et al. (2024) claimed their RAG framework is "reliable"—this conclusion is based primarily on a small-scale knowledge base ($\sim$100 papers) and a minimal qualitative evaluation (3 Q\&A pairs). This "Proof-of-Concept" (PoC) approach obscures a critical question: When the knowledge to be injected expands to the entire domain's hundreds of thousands of papers, where is the "performance ceiling" of naive RAG? Is it still "reliable"?

\begin{figure}[h!]
    \centering
    \includegraphics[width=0.4\textwidth]{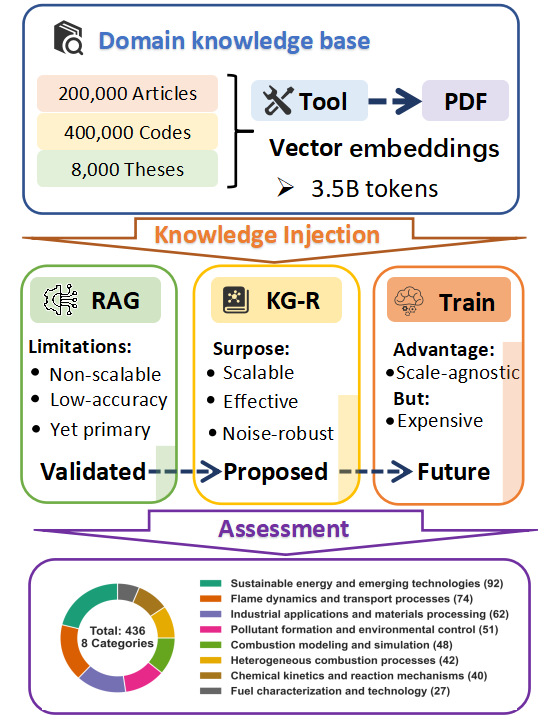}
    \caption{Our proposed three-stage domain knowledge injection framework. It integrates (Top) a knowledge base, (Bottom) an evaluation benchmark, and (Middle) an evolution roadmap (Validated).}
    \label{fig:overall_structure}
\end{figure}

Based on the premise, this paper proposes a robust knowledge-injection framework for building a foundation model in combustion science, with the core architecture shown in Fig.~\ref{fig:overall_structure}. The framework comprises: (1) the first “AI-ready” combustion knowledge base at the 3.5 billion-token scale, constructed from multimodal sources including large bodies of scholarly articles and experimental knowledge base (Top); (2) \textit{CombustionQA}, an objective and comprehensive evaluation benchmark with 436 questions covering major subareas of combustion (bottom); and (3) a three-stage evolution path for knowledge injection, spanning naive RAG, knowledge-graph-enhanced RAG, and domain fine-tuning. Using this framework, we validate the first stage (naive RAG) and demonstrate that more advanced injection routes are necessary for comprehensive domain coverage.

% The core work of this proposal is to quantitatively "validate" the first stage of this framework (naive RAG). Our empirical results reveal its severe performance and scalability ceilings ($\sim$3k document bottleneck). This quantitative evidence demonstrates for the first time that Stage 1 is insufficient for large-scale scientific knowledge. Therefore, our framework provides the combustion AI community with a much-needed "yardstick" and "roadmap," indicating the \textbf{"necessity"} of evolving towards Stage 2: KG-RAG (Proposed) and Stage 3: Domain Fine-Tuning (Future).

% -------------------------------------------------------------------- %
% --- Section 2: Methodology ---
% -------------------------------------------------------------------- %

\section{Framework Methodology\label{sec:methodology}}
\addvspace{10pt}

% Our three-stage knowledge injection framework (Fig.~\ref{fig:overall_structure}) is built upon two core, foundational assets that we independently constructed: (1) a large-scale, "AI-Ready" knowledge base; and (2) a rigorous, graduate-level evaluation benchmark.

\subsection{Knowledge Base Construction\label{sec:knowledge_datasets}}

A reliable, comprehensive knowledge base is the essential substrate for a domain foundation model. In this framework, we construct a multimodal knowledge base spanning the full breadth of combustion knowledge, including peer-reviewed literature, theses and dissertations, experimental and simulation datasets, and combustion CFD code. To render these materials AI-ready, we employ established parsing tools to extract tokens and structure from raw sources. For “dark data” PDFs, the pipeline performs layout parsing and metadata extraction, identifying titles, authors, body text, figures, tables, and captions, then produces vector embeddings. The pipeline also handles specialized extraction of equations, reaction mechanisms, units, and tabular data while preserving provenance.

\begin{figure}[h!]
    \centering
    \includegraphics[width=172pt]{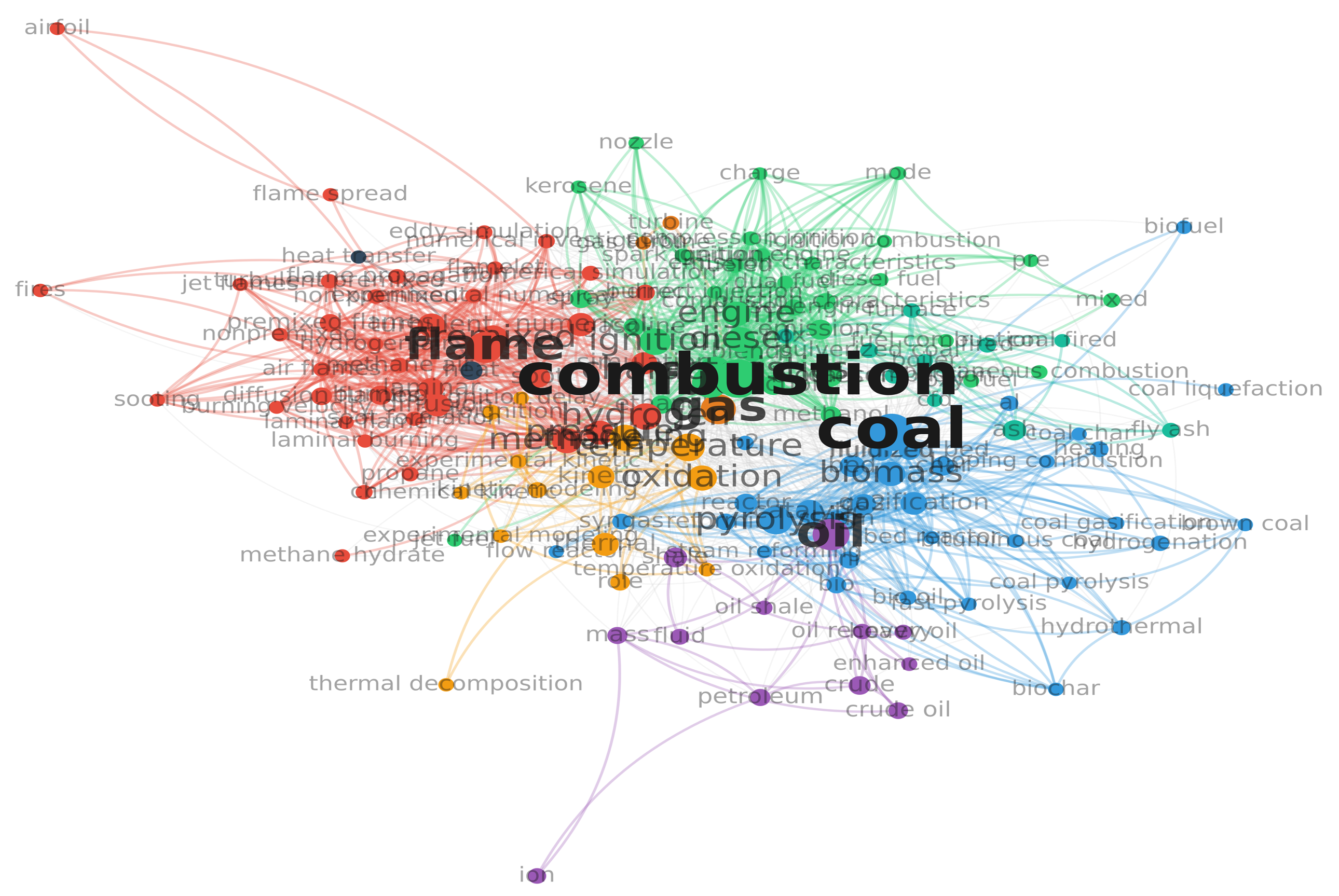} % 请替换为您实际的图2文件名
    \caption{\footnotesize Knowledge landscape of combustion science derived from the proposed knowledge base.} %The clusters clearly reproduce the major branches of combustion science (e.g., chemical kinetics, turbulent combustion) in the semantic space, demonstrating the high effectiveness of our parsing and vectorization.}
    \label{fig:knowledge_cluster}
\end{figure}

The current release aggregates approximately 200{,}000 peer-reviewed articles, approaching the cumulative output of mainstream combustion journals, together with 8{,}000 theses and dissertations and roughly 400{,}000 lines of combustion CFD code. Fig.~\ref{fig:knowledge_cluster} visualizes topic coverage of the textual subset via a keyword co-occurrence network, indicating breadth across flames, ignition and oxidation chemistry, reactors, engines, and related subareas, yielding near-comprehensive domain coverage. These materials comprise approximately 3.5~billion tokens, sufficient for a wide range of domain model injection strategies.

% Constructing an "AI-Ready" specialized domain knowledge base is an extremely arduous knowledge engineering task. Our goal was to transform 100,000 "dark data" PDFs into a precise, reliable, and retrievable Domain Knowledge Vector Base.

% \subsubsection{Corpus Scale and Curation}
% Our knowledge base (Phase I) comprises 100,000 documents, covering journal papers (e.g., \textit{PECS}, \textit{C\&F}, \textit{PNF}), conference papers, review papers, and technical reports. Sources include core databases like Web of Science, Scopus, Springer, and Elsevier, as well as private datasets provided by domain experts.

% \subsubsection{The Uni-Parser Pipeline}
% We deployed a "Uni-Parser" pipeline to ensure knowledge fidelity:
% \begin{enumerate}
%     \item \textbf{Metadata Parsing:} Automatically extract title, authors, and year from filenames, DOIs, and CrossRef.
%     \item \textbf{Structural Parsing:} Automatically identify and separate title, abstract, body text, figures/tables, and references, reconstructing the document's original section tree.
%     \item \textbf{Triple-Storage:} Process each document into three forms: raw text chunks, table/figure data, and semantic embedding vectors.
% \end{enumerate}

% \subsubsection{Semantic Validation}
% To validate the semantic quality of this "AI-Ready" knowledge base, we performed unsupervised clustering.

% This validated 100,000-document knowledge base (Fig.~\ref{fig:knowledge_cluster}) lays an extremely solid foundation not only for our current RAG evaluation but also for future work, such as knowledge graphs (KGs).

\subsection{Benchmark Construction: \textit{CombustionQA}\label{sec:benchmark_construction}}

\begin{figure}[h!]
    \centering
    \includegraphics[width=192pt]{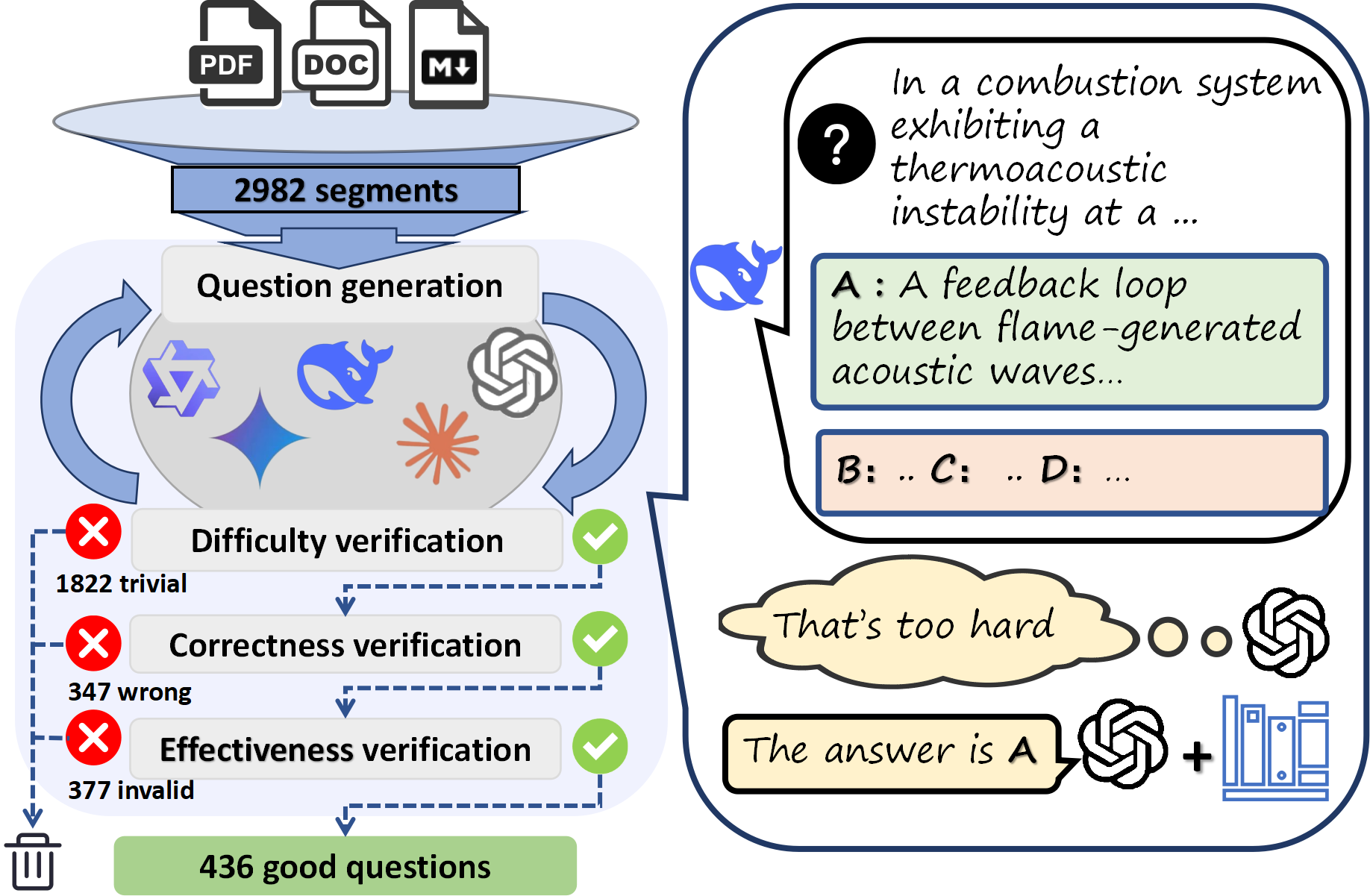} % 请替换为您实际的图3文件名
    \caption{\footnotesize The multi-stage construction process for \textit{CombustionQA} (436 questions).}
    \label{fig:qa_pipeline}
\end{figure}

Just as combustion diagnostics require calibration, a domain LLM also requires objective, domain-grounded standards for evaluating its capabilities. We curate \textit{CombustionQA} through a rigorous generation and validation pipeline, as illustrated in Fig.~\ref{fig:qa_pipeline}.

First, 2,982 high-information segments are selected from peer-review articles and comprehensive reports within the corpus. These segments serve as the foundation for an agentic question generation and validation system, comprising two automated verification stages followed by manual curation.

In the difficulty filtering stage, an agentic system generates candidate questions from source segments, with each question accompanied by a model-generated answer, explanation, and citation of supporting sentences. GPT-5 then attempts to answer these questions in a zero-shot setting without source context. Questions answered correctly are returned to the generation model for refinement with increased difficulty, as illustrated in the right panel of Fig.~\ref{fig:qa_pipeline}. Segments that fail to produce sufficiently challenging questions after five bootstrapping iterations are discarded; 1,822 segments are eliminated through this process.

In the correctness verification stage, retained questions are validated by providing GPT-5 with the cited source sentences. Questions must be answerable given ideal context; those that fail this validation—indicating ambiguity or insufficient information—are discarded,as shown in left panel of Fig.~\ref{fig:qa_pipeline}, removing 347 invalid cases.

Finally, manual curation eliminates questions with incomplete information, temporal dependencies, or other quality issues, removing an additional 377 cases. The resulting 436-question benchmark spans eight core combustion subfields (Fig.~\ref{fig:overall_structure}).

\subsection{Injection \& Evaluation Protocol\label{sec:evaluation_protocol}}

Our framework defines a three-stage injection path spanning naive RAG, knowledge-graph-enhanced RAG, and domain fine-tuning, as shown in the middle of Fig.~\ref{fig:overall_structure}. Existing knowledge injection methods form a resource spectrum: RAG represents the lightweight end, requiring minimal resources, while model training approaches~\cite{Hu.2021,Ouyang.2022,Gururangan.2020} demand substantial computational resources but are believed to confer more fundamental capabilities by internalizing knowledge into model weights. Knowledge-graph-enhanced RAG (KG-RAG)~\cite{edge2025localglobalgraphrag} offers an intermediate approach, incorporating structured domain knowledge to enhance retrieval precision while maintaining computational efficiency.

This work validates the first stage through controlled experiments. Our implementation employs FAISS~\cite{johnson2019billion} for vector storage, BGE-M3~\cite{chen2024bge} for embeddings, and LangChain~\cite{chase2022langchain} for orchestration. Given a \textit{CombustionQA} query, the system retrieves the top-5 most relevant chunks via cosine similarity and feeds them with the query into the LLM. The framework enables controlled retrieval scope variation to diagnose retrieval quality and context-contamination effects.

We designed controlled experiments to quantify performance and core bottlenecks under controlled retrieval conditions. Detailed experimental design and results are presented in Section~\ref{sec:results}.

%Based on the "Knowledge Base" (Sec.~\ref{sec:knowledge_datasets}) and "CombustionQA Benchmark" (Sec.~\ref{sec:benchmark_construction}), our framework (Fig.~\ref{fig:overall_structure}) defines a three-stage injection path. This work focuses on the quantitative validation of the framework's first stage: Naive RAG.

%For domain experts, the RAG principle is embodied in our framework (Fig.~\ref{fig:overall_structure}) as a dynamic process:
%A \textbf{Query} from the "Evaluation Benchmark" (Fig.~\ref{fig:overall_structure}, Bottom) is posed;
%The system performs semantic \textbf{Retrieval} in the "Knowledge Base" (Fig.~\ref{fig:overall_structure}, Top) to fetch relevant context;
%Finally, this context, along with the original query, is fed into the LLM (the "vRAG" module, Fig.~\ref{fig:overall_structure}, Middle) for \textbf{Augment \& Generate} to produce the final answer.

%To this end, we designed a set of rigorous control experiments to quantitatively diagnose the performance, bottlenecks, and costs of this process. The detailed experimental design and results are presented in Section~\ref{sec:results}.

\section{Results and Findings\label{sec:results}}
\addvspace{10pt}

\subsection{The Performance Ceiling of Naive RAG\label{sec:optimal_ceiling}}
\begin{figure}[h!]
    \centering
    \includegraphics[width=192pt]{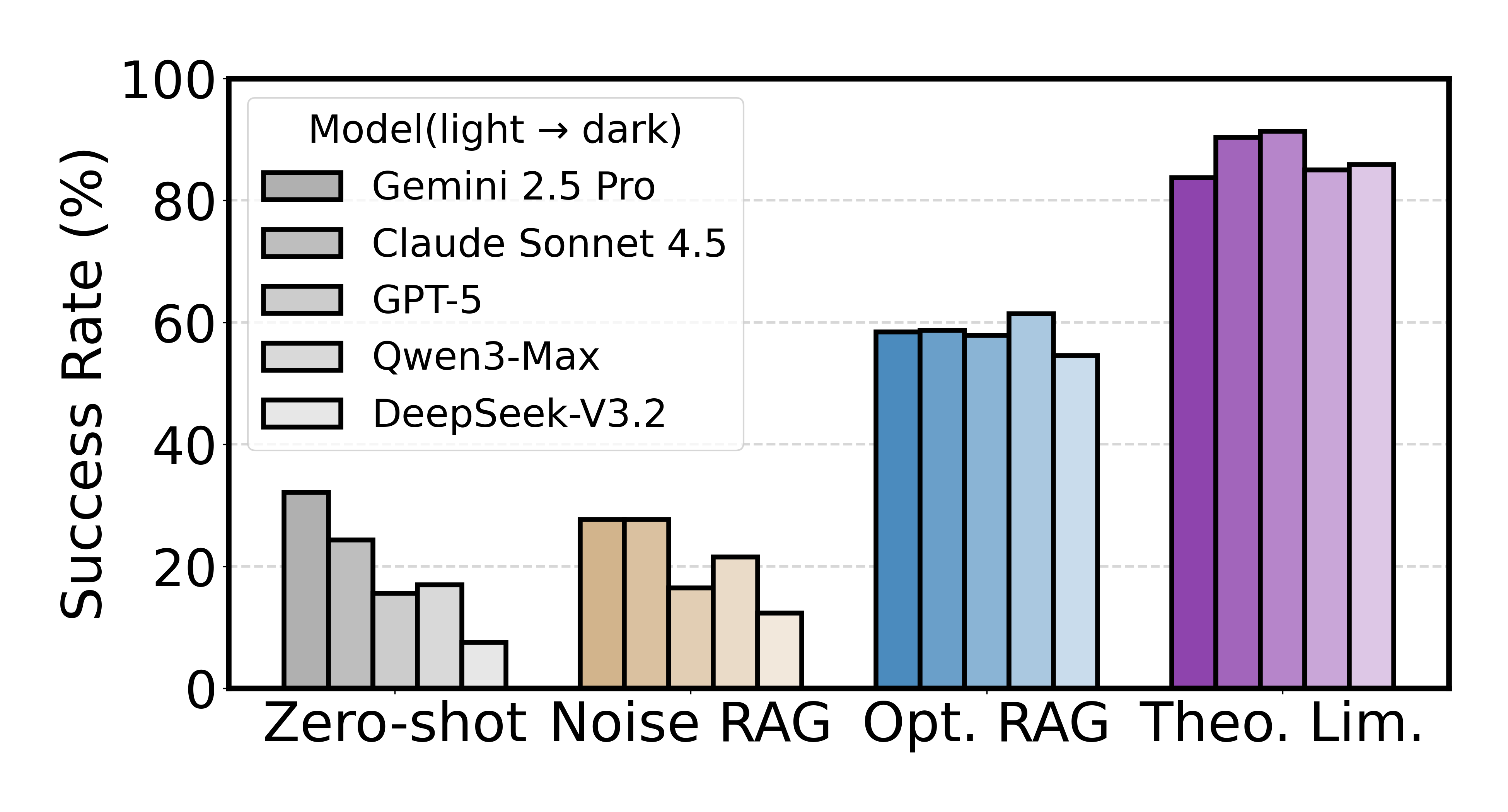} % 请替换为您实际的图4文件名
    \caption{\footnotesize Accuracy of the four core scenarios across different LLMs.}
    \label{fig:four_scenarios}
\end{figure}

To quantify the upper bound of naive RAG performance, we designed four controlled evaluation scenarios as shown in Table~\ref{tab:scenarios}: Zero-Shot, where the LLM answers without any retrieved context; Theoretical Upper Bound, where the LLM is directly provided with the source citation; Optimal RAG, where retrieval is performed within a perfectly condensed knowledge base of 2,982 source segments; and Noise RAG, where retrieval targets an equivalent volume of high-quality but non-answer domain content.
Fig.~\ref{fig:four_scenarios} presents these scenarios ordered by ascending performance across multiple mainstream LLMs.

\begin{table}[h!]
    \centering
    \footnotesize % 缩小字体以适应双栏
    \caption{Feature matrix of the four evaluation scenarios. *``Noise" is defined as high-quality distracting information irrelevant to the answer.}
    \label{tab:scenarios}
    \begin{tabular}{|l|c|c|c|c}
        \hline
        \textbf{Scenario} & \textbf{\shortstack{Provides \\ Context}} & \textbf{\shortstack{Contains \\Answer}} & \textbf{\shortstack{Contains \\ Noise*}} \\
        \hline
        Zero-Shot  & No & N/A  & N/A  \\
        Theoretical Limit& Yes & Yes & No \\
        Optimal RAG & Yes & Yes & Yes \\
        Noise RAG & Yes & No & Yes \\
        \hline
    \end{tabular}
\end{table}

Three key findings emerge from these results. First, the 60+ percentage-point gap between Theoretical Upper Bound (87.3\%) and Zero-Shot (23.35\%) confirms that \textit{CombustionQA} questions are challenging yet answerable. Second, Noise RAG (21.1\%) performs worse than Zero-Shot, demonstrating that retrieval of irrelevant content actively degrades performance—bad RAG is worse than no RAG. Third, and most critically, even Optimal RAG achieves only 58.24\% accuracy, falling nearly 30 percentage points short of the theoretical upper bound. This performance gap under ideal retrieval conditions reveals fundamental limitations in the naive RAG architecture for comprehensive domain knowledge integration.

\subsection{Bottleneck Analysis: Diagnosing the 30\% Performance Gap\label{sec:3.2}}
To understand why Optimal RAG falls 30 percentage points short of the theoretical ceiling, we analyzed retrieval outcomes and their corresponding accuracy.

\begin{figure}[h!]
    \centering
    \includegraphics[width=192pt]{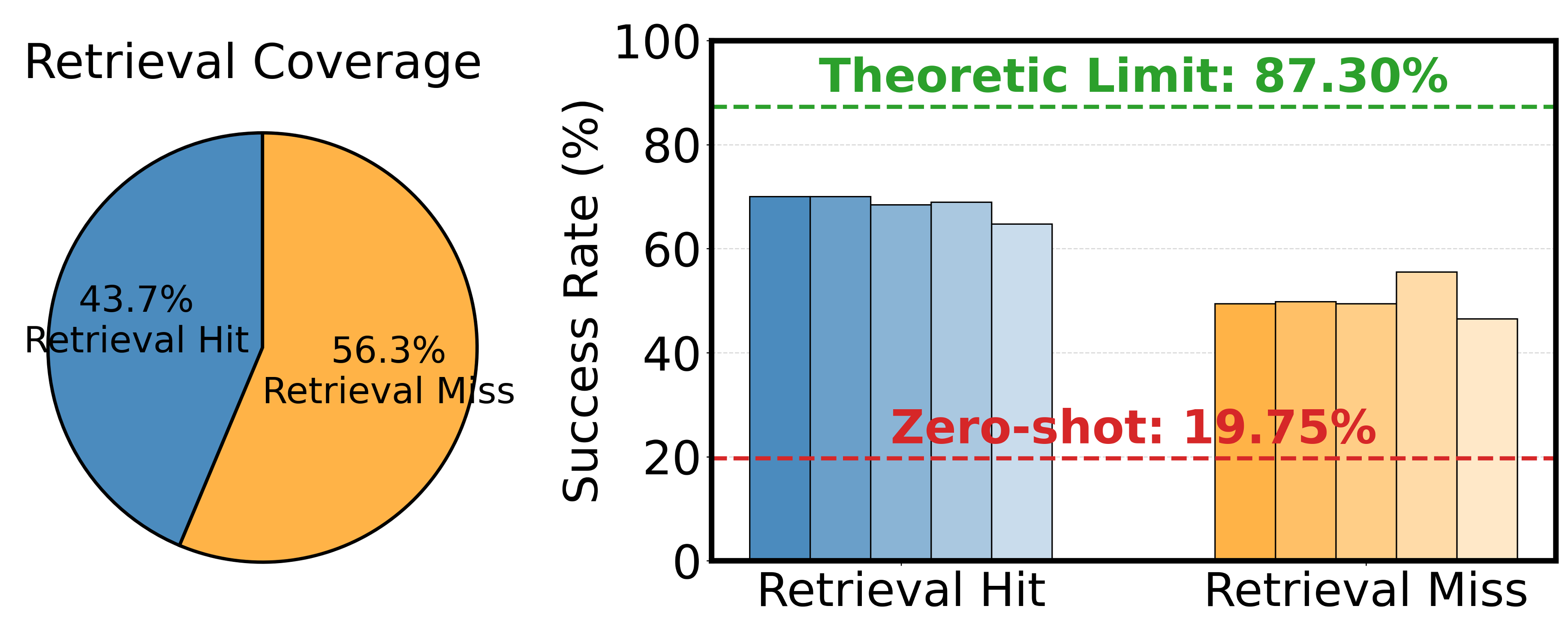} % 请替换为您实际的图5文件名
    \caption{\footnotesize Performance comparison under ``Optimal RAG" scenario, categorized by whether the exact source segment was retrieved (Hit vs. Miss).}
    \label{fig:bottleneck_analysis}
\end{figure}
% 这里图例我没想到什么好办法，之后再琢磨琢磨

Fig.~\ref{fig:bottleneck_analysis} shows that across all questions, 43.7\% successfully retrieved the source segment (Hit), while 56.3\% failed (Miss). Performance diverges sharply between these two groups: Hit cases achieve approximately 70\% accuracy, while Miss cases drop to approximately 50\%.

This reveals two distinct bottlenecks limiting Optimal RAG performance. First, retrieval recall is insufficient: even within a perfectly condensed 2,982 segment corpus containing all answer sources, the retriever misses the correct segment in 56.3\% of cases, a failure rate that will intensify as corpus scale increases. Second, context contamination degrades answer quality: even in Hit cases, accuracy remains 17 percentage points below the Theoretical Upper Bound.  This gap arises because retrieval returns multiple segments: although the source segment is present, other retrieved content acts as semantic noise that obscures the model's ability to derive the correct answer.

Taken together, the controlled experiments in this section provide an empirical anchor for Stage~1 of our framework: they establish naive RAG's performance ceiling and identify its two core failure modes---retrieval miss and context contamination. Addressing these failures through structured domain knowledge (Stage~2: KG-RAG) and weight-level knowledge internalization (Stage~3: continued pretraining) constitutes the immediate next step of this research program.

% -------------------------------------------------------------------- %
% --- Section 4: Conclusion ---
% -------------------------------------------------------------------- %

\section{Conclusion and Future Work\label{sec:conclusion}}
\addvspace{10pt}

This work delivers three contributions toward domain foundation models for combustion science.

First, we construct essential community infrastructure: an AI-ready multimodal knowledge base aggregating approximately 200{,}000 peer-reviewed articles, 8{,}000 theses and dissertations, and 400{,}000 lines of combustion CFD code, together with \textit{CombustionQA}, an evaluation benchmark of 436 questions spanning eight core subfields.

Second, using this infrastructure we provide the first controlled quantification of naive RAG in the combustion domain. Even under ideal retrieval conditions, accuracy peaks at 58\%---nearly 30 percentage points below the theoretical ceiling (87\%)---while irrelevant retrieval actively degrades performance below the zero-shot baseline. Diagnostic analysis attributes this gap to two bottlenecks: a 56\% retrieval miss rate and context contamination that suppresses accuracy even when the correct source is retrieved.

Third, these findings yield an actionable roadmap. Because the identified bottlenecks are structural to vector-similarity retrieval, overcoming them requires either knowledge-graph-enhanced retrieval (Stage~2) to impose structured supervision on context selection, or continued pretraining (Stage~3) to internalize domain knowledge into model parameters. The framework and benchmarks established here will serve as shared infrastructure for validating these next-stage strategies.

% -------------------------------------------------------------------- %
% --- Acknowledgments ---
% -------------------------------------------------------------------- %

\acknowledgement{Declaration of competing interest}
\addvspace{10pt}

There is no competing interest.

% \acknowledgement{Acknowledgments}
% \addvspace{10pt}

% Funding for ZL was provided by... We thank [Collaborators] for their insights on...

% -------------------------------------------------------------------- %
% --- References ---
% -------------------------------------------------------------------- %

\footnotesize
\baselineskip 9pt
\clearpage

\bibliographystyle{pci}
\bibliography{PCI_LaTeX}

\end{document}